%% file: main.tex
\pdfoutput=1

\documentclass[10pt,twocolumn,letterpaper]{article}

\usepackage{iccv}              

\input{preamble}

\definecolor{iccvblue}{rgb}{0.21,0.49,0.74}
\usepackage[pagebackref,breaklinks,colorlinks,allcolors=iccvblue]{hyperref}

\def\paperID{*****} 
\def\confName{CVPR}
\def\confYear{2026}

\title{Hardware-Aware Neural Feature Extraction for Resource-Constrained Devices}

\author{
Francesco Tosini$^{1}$ \and
Simone Pedroni$^{1}$ \and
Christian Veronesi$^{1}$ \and
Pietro Bartoli$^{1}$ \and
Andrea Giudici$^{1}$ \and
Marco Paracchini$^{1}$ \and
Marco Marcon$^{1}$ \and
Diana Trojaniello$^{2}$ \\
$^{1}$Department of Electronics, Information and Bioengineering (DEIB), Politecnico di Milano, Italy\\
$^{2}$Smart Eyewear Lab, EssilorLuxottica, Milan, Italy\\
{\tt\small francesco.tosini@polimi.it}
}

\makeatletter
\renewcommand{\@maketitle}{%
  \newpage
  \null
  \vspace{-0.45in}
  \begin{center}
    {\large\color{gray}
    This paper has been accepted for publication at the\\
    IEEE Conference on Computer Vision and Pattern Recognition (CVPR) Workshops, 2026. ©IEEE
    \par}
    
    \vspace{0.28in}
    
    {\Large\bfseries \@title \par}
    
    \vspace{0.22in}
    
    {\large
    \lineskip .5em
    \begin{tabular}[t]{c}
    \@author
    \end{tabular}
    \par}
  \end{center}
  \par
  \vspace{0.20in}
}
\makeatother

\begin{document}
\maketitle
\input{sec/0_abstract}    
\input{sec/1_intro}
\input{sec/2_methodology}
\input{sec/3_results}
\input{sec/4_conclusion}
{
    \small
    \bibliographystyle{ieeenat_fullname}
    \bibliography{main}
}

\end{document}

%% file: preamble.tex
\usepackage{multirow}
\usepackage{makecell}

%% file: sec/0_abstract.tex
\begin{abstract}
Visual SLAM is a core component of spatial computing systems, yet deploying learned local feature extractors on microcontroller-class hardware remains challenging due to memory, bandwidth, and quantization constraints. While modern neural descriptors provide strong robustness, their practical adoption is often hindered by system-level bottlenecks that are not captured by FLOP-based efficiency metrics.
In this work, we introduce \textit{Gideon}, a hardware-aware neural feature extractor explicitly designed for resource-constrained devices. Our approach combines relational knowledge distillation from a SuperPoint teacher with differentiable neural architecture search (DNAS) under strict memory and operator constraints. Unlike conventional design pipelines, we treat quantization stability and dynamic-range compactness as first-class objectives.
We show that architectural choices such as replacing Batch Normalization with Affine layers significantly improve INT8 robustness, and that descriptor dimensionality directly governs quantization resilience. Deployed on STM32N6, Gideon achieves 9.003\,ms inference time (111\,fps) while remaining below 1.5\,MB memory footprint. Remarkably, INT8 quantization induces negligible degradation and occasionally matches full-precision performance.
These results demonstrate that robust learned feature extraction can be reconciled with embedded hardware constraints through holistic hardware–algorithm co-design.
\end{abstract}

%% file: sec/1_intro.tex
\section{Introduction}
\label{sec:intro}

Despite living in an era of unprecedented technological advancement, our interface with the digital world remains paradoxically primitive. In fact, we are still relying on small 2D rectangles to experience the virtual world, which causes a significant amount of friction. Similarly to the transition from command-line interfaces to graphical user interfaces, we are experiencing a paradigm shift led by the introduction of Spatial Computing. Smart Eyewear, in particular, stands at the forefront of this revolution with its promise to weave the virtual domain directly into our physical perception. However, this requires solving several engineering challenges, including the need for robust, real-time machine perception on a device with extreme thermal, energetic, and usability constraints.

Visual Simultaneous Localization and Mapping (SLAM) is at the core of the Smart Glasses' perception stack: it provides the world knowledge required to anchor virtual content to the real world, understanding ego-motion and the 3D structure of the environment.

Modern SLAM algorithms, such as ORBSLAM-3 \cite{orbslam3}, rely heavily on the extraction of visual features, i.e. distinctive keypoints and their associated descriptors, which serve as the fundamental "anchors" for tracking. Ideally, these features must be repeatable across drastic changes in viewpoint and illumination, yet cheap enough to compute at high frame rates.

The current State of the Art presents a sharp dichotomy. On one hand, traditional "hand-crafted" feature extractors (such as ORB \cite{orb}, FAST \cite{fast}, or BRISK \cite{brisk}) are computationally efficient and have served as the backbone of SLAM for years. Yet, they struggle to stay reliable in challenging scenarios, such as texture-less environments or with dynamic lighting. On the other hand, deep learning models such as SuperPoint \cite{superpoint} and R2D2 \cite{r2d2}, as well as methods such as D2-Net~\cite{Dusmanu2019CVPR} and DISK~\cite{NEURIPS2020_a42a596f}, managed to reframe the feature extraction process as a fully differentiable problem, producing “learned” features that offer unprecedented robustness and repeatability. However, this performance comes at a prohibitive cost on wearable devices: these models typically rely on heavy backbones that far exceed the budget of low-power edge neural processing units (NPUs), which becomes especially true for newer transformer-based approaches, such as LoFTR \cite{loftr} and LightGlue~\cite{lindenberger2023lightglue}. In fact, deploying a full-scale feature extraction model on Smart Eyewear would result in unacceptable latency and rapid battery depletion.

The State of the Art offers lightweight architectures such as XFeat-Micro~\cite{xfeat} and ALIKED-Tiny~\cite{aliked}. However, despite their theoretical efficiency, these models are often not well-suited for microcontroller-class devices due to system-level bottlenecks not captured by FLOP-based metrics. 
In particular, semi-dense or high-dimensional descriptors introduce significant data movement overhead, saturating memory bandwidth, while operations such as dilated or deformable convolutions lead to irregular memory access patterns that degrade cache locality. 
As a result, these architectures, although effective on edge GPUs and mobile NPUs, remain difficult to deploy on MCUs without substantial architectural modifications.

Gideon circumvents these architectural pitfalls through strict hardware-algorithm co-design. By favoring a topology comprised of standard convolutional patterns, the model ensures contiguous memory access and a highly predictable data flow.

This design choice maximizes cache hit rates and eliminates the overhead associated with sparse operations. Consequently, Gideon keeps its entire execution footprint (both weights and activations) strictly under the SRAM threshold, allowing it to fully leverage the NPU's acceleration capabilities without triggering bandwidth or memory bottlenecks. 

Beyond lightweight feature extractors, recent works have demonstrated fully onboard SLAM and visual–inertial odometry on highly resource-constrained platforms. 
For instance, NanoSLAM~\cite{nanoSLAM2309.12008} enables complete SLAM pipelines on ultra-low-power processors, while LEVIO~\cite{levio2602.03294} proposes a lightweight visual–inertial odometry system tailored for embedded devices. 
These efforts highlight the feasibility of real-time perception under strict power and memory budgets.

However, such systems typically rely on handcrafted or fixed feature extraction modules and do not explicitly address quantization-aware learning of local features under microcontroller-class constraints. 
In contrast, our work focuses on the hardware-aware learning of compact and quantization-stable feature representations, which can serve as drop-in components within broader SLAM pipelines and enable distributed edge-SLAM paradigms where feature extraction runs on constrained endpoints. While full SLAM integration is beyond the scope of this work, we report feature-level metrics such as repeatability and matching correctness, which are widely accepted indicators of downstream SLAM performance.

In this work, we bridge this gap between the robustness of learned features and the efficiency of hand-crafted ones with Gideon, a novel training paradigm and neural architecture specifically designed for the intrinsic limitations of wearable devices. 

Rather than training from scratch, our approach relies on Knowledge Distillation to transfer the robust "intuition" of a SuperPoint teacher into a lightweight student. Instead of manually designing the network, which is expensive and error-prone, we employ Differentiable Network Architecture Search (DNAS) to automatically identify an efficient topology that balances accuracy and cost. This approach accounts for the tight constraints of wearable devices, allowing us to deploy robust neural perception at the edge. 

Unlike prior approaches that optimize network architecture, distillation, and deployment independently, we explicitly formulate feature extraction as a constrained system-level problem. In our setting, memory footprint, dataflow regularity, and quantization stability are treated as primary optimization objectives during training, rather than post-hoc constraints. 

%% file: sec/2_methodology.tex
\section{Methodology}
\label{sec:methodology}

In this section, we describe the design and training strategy underlying \textit{Gideon}. 
Rather than treating architecture, distillation, and deployment as independent components, we adopt a unified hardware-aware perspective in which model topology, loss formulation, and optimization are jointly shaped by embedded system constraints.

\begin{figure*}[th]
  \centering
  \includegraphics[width=1.0\linewidth]{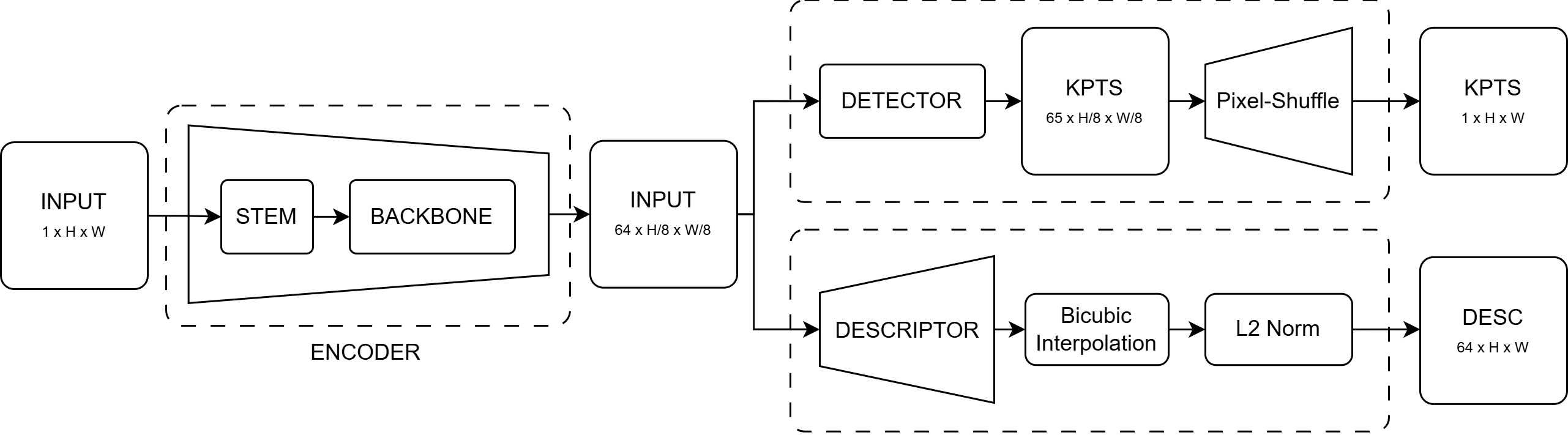}
  \caption{Overview of the baseline functional topology, inspired by the original SuperPoint \cite{superpoint}.}
   \label{fig:func_topology}
\end{figure*}

\subsection{Hardware-Aware Design Principles}
\label{sec:hardware_principles}

The design of \textit{Gideon} is driven by system-level constraints rather than purely architectural preferences. 
On microcontroller-class devices, latency and robustness are dominated not only by model size or FLOPs, but by memory footprint, data movement, and quantization stability.

First, the limited on-chip SRAM budget constrains both weights and intermediate activations, requiring compact feature maps and strictly bounded descriptor dimensionality to ensure full in-memory execution. 
Second, irregular memory access patterns can severely degrade throughput on embedded NPUs; therefore, we restrict the architecture to standard convolutional operators with predictable and contiguous dataflow. 
Third, INT8 inference imposes tight constraints on activation dynamic range, making quantization stability a primary design objective rather than a post-training adjustment.

These considerations redefine the optimization target from raw representational capacity to \emph{system-level density}, where robustness, dynamic-range compactness, and hardware compatibility are jointly optimized.

\subsection{Topological Structure}

Guided by the hardware constraints outlined in Section~\ref{sec:hardware_principles}, we established a baseline functional architecture represented in Fig. \ref{fig:func_topology}, composed of a single shared encoder and two task-specific decoder heads similarly to the original SuperPoint.

The shared encoder is made up of a fixed stem and a searchable architectural block, which is dynamically determined by the DNAS pipeline. This improves search efficiency and results in better models. 
This bipartite encoder processes the input image to extract a dense latent representation, which is then fed into two parallel branches: a detector head, which computes a full-resolution spatial probability map of interest points through pixel-shuffling, and a descriptor head, which generates a semi-dense map of L2-normalized, high-dimensional feature vectors. This design allows the network to optimize both keypoint localization and description, while reducing the computational cost of the backbone.

\subsection{Knowledge Distillation and NAS}

Training a compact model from scratch under strict parameter and memory budgets often leads to sub-optimal feature representations. 
To overcome this limitation, we adopt a knowledge distillation strategy in which the student network, \textit{Gideon}, is trained to mimic the activation manifold of a frozen SuperPoint teacher~\cite{hinton2015distilling,DBLP:journals/corr/RomeroBKCGB14}. 
Rather than rediscovering robust features from raw supervision alone, the student learns to map an already structured feature space, enabling efficient compression of the teacher's representational power.

However, distillation alone does not guarantee hardware efficiency. 
Given the stringent latency, memory, and quantization constraints of embedded NPUs, the student architecture itself must be explicitly optimized for deployment. 
Manual design under such constraints proved unstable and suboptimal, particularly for shallow networks operating near the edge of their capacity.

To jointly address representational fidelity and hardware compatibility, we introduce a Differentiable Neural Architecture Search (DNAS) stage inspired by DARTS~\cite{darts}. 
During the search phase, a super-network (``SuperGideon'') replaces each architectural block with a stochastic mixture of candidate operations (e.g., residual bottlenecks, Inception-like modules, standard convolutions). 
Both network weights and architectural parameters are optimized end-to-end.

The discrete operator selection is relaxed through the Gumbel-Softmax reparameterization~\cite{gs1,gs2}, enabling differentiable exploration of the search space. 
As the temperature is annealed, the architecture progressively converges toward a deterministic topology that satisfies hardware constraints while preserving the relational structure distilled from the teacher.

The search space includes standard convolutional blocks and Residual, Bottleneck and Inception-like modules with varying kernel sizes. Hardware constraints are implicitly enforced through architectural priors, rather than explicit latency terms, ensuring compatibility with MCU deployment.

\subsection{Loss Design}

Initially, training relied on MSE to match teacher outputs, which proved suboptimal as it constrains the student to replicate absolute activations rather than optimizing for downstream performance.



Furthermore, since the network performs detection and description in parallel, effectively weighting the individual contribution of each task is crucial. To ensure a stable and balanced training process, we employed uncertainty-based weighting to dynamically scale the respective loss functions.

\subsubsection{Detection Loss}
\label{sec:detection_loss}

The number of background pixels vastly outnumbers the keypoints, which are ideally sparse peaks in the signal. 
As a result, standard losses such as Binary Cross-Entropy or Mean Squared Error tend to bias the model toward trivial background predictions and penalize near-miss activations as false positives.

To overcome these issues, we adopt a variant of the Focal Loss originally proposed in CornerNet. 
In addition to reweighting hard examples, we assign soft labels to the neighborhood of each keypoint using a 2D Gaussian kernel, reducing the penalty for spatially close predictions while preserving strong gradients for true positives.
This dual mechanism, spatial penalty reduction and dynamic confidence scaling, encourages the optimization process to focus on hard negatives and precise center localization, allowing the model to produce sharper, better-localized heatmaps while preventing the background signal from dominating the gradient.

Furthermore, SuperPoint typically operates with a notably low detection threshold (0.005), which often correlates with noisy predictions. In contrast, by anchoring our optimization to binarized ground truth locations, even while smoothing the loss surface with the aforementioned Gaussian penalty, the network is forced to suppress ambiguity. This has proven to yield significantly sharper activation peaks, resulting in higher detection confidence and accuracy.

To further benefit from this, we have introduced an Adaptive Thresholding mechanism that dynamically adjusts the detection sensitivity during training by computing the exponential moving average (EMA) of the average activation of the top pixels. This introduces a soft constraint on the mean number of keypoints and ensures that the extracted features are robust.

The implicit target number of keypoints is not fixed a priori, but emerges from the EMA dynamics of high-confidence activations. In practice, this results in stable keypoint densities across diverse scenes, including both indoor and outdoor environments.

\subsubsection{Descriptor Loss}
\label{sec:descriptor_loss}

SuperPoint descriptors feature low variance, suggesting that most of their information is either redundant or unnecessary. Moreover, forcing a compact student network to perfectly replicate the absolute, high-dimensional metric space of the teacher often limits its representational flexibility and can negatively affect convergence.

To address this, we departed from standard Euclidean minimization, shifting the distillation objective from absolute feature matching to relational matching. Consequently, the student is trained to replicate the internal geometric relationships, i.e. the similarity manifold, of the teacher's feature space rather than attempting to align individual descriptors.

This works by computing a dense self-similarity matrix through the cosine similarity between descriptors across all spatial locations for both the student and the frozen teacher. These correlation matrices are then converted into probability distributions over the spatial dimensions using a temperature-scaled softmax operation. This yields a probabilistic map representing how every local feature relates to all other features within the image.

Finally, the student is optimized to match the teacher's relational distribution by minimizing the Kullback-Leibler (KL) Divergence~\cite{kullback1951information}:

\begin{equation}
    \mathcal{L}_{desc} = \frac{1}{N} \sum_{i=1}^{N} \text{KL} \left( \;\sigma \left( \frac{S^{gt}_i}{\tau} \right) \;\Bigg\| \; \sigma \left( \frac{S^{pred}_i}{\tau} \right) \; \right)
\end{equation}

where $S^{gt}_i$ and $S^{pred}_i$ denote the dense self-similarity matrices for the $i$-th spatial location of the teacher and the student respectively, $\sigma$ represents the softmax operation, $\tau$ is a temperature scaling parameter controlling the sharpness of the probability distribution, and $N$ is the total number of spatial locations. 

This approach allows the student to construct its own highly efficient embedding space, provided it faithfully preserves the same structural correlation topology as the teacher.


\subsection{Training Procedure}

\subsubsection{Dataset and Data Augmentation}

The proposed model is trained using $29000$ images from the TUM-VI dataset \cite{tumvi} ($23200$ for training, $5800$ for validation and test purposes). Input images are initially resized to $256 \times 256$ pixels and subsequently center-cropped to a spatial resolution of $192 \times 256$. To enhance the network's rotational and viewpoint invariance without incurring CPU bottlenecks, we employ dynamic, on-the-fly geometric data augmentations executed directly on the GPU. During each training iteration, inputs and their corresponding pseudo-ground truth targets undergo random spatial transformations, including $90^\circ$, $180^\circ$, and $270^\circ$ rotations, as well as horizontal and vertical flips.

\subsubsection{Optimization and Scheduling}

The network parameters are optimized using the AdamW optimizer with an initial learning rate of $1 \times 10^{-3}$ and a weight decay of $1 \times 10^{-4}$. To ensure training stability and mitigate the risk of exploding gradients, gradient clipping is enforced with a maximum norm of $5.0$. The learning rate is dynamically adjusted using a \textit{ReduceLROnPlateau} scheduler, which monitors the validation loss and reduces the learning rate by a factor of $0.5$ if no improvement is observed for a patience of $5$ epochs. The two loss functions described in Sections \ref{sec:detection_loss} and \ref{sec:descriptor_loss} are dynamically balanced via uncertainty weighting \cite{uncertaintyweighting} during training, whereas they are simply summed for validation purposes.

\subsection{Inference Protocol}
\label{sec:inference_protocol}

During training, the network predictions are used directly without Non-Maximum Suppression (NMS), and losses are computed densely on pixel-shuffled logits and $L_2$-normalized descriptors. 
Teacher heatmaps are filtered with NMS (spatial radius $=4$) and thresholded at $0.005$.

At inference time, predicted heatmaps undergo spatial NMS ($r=4$) followed by adaptive thresholding (Section~\ref{sec:detection_loss}) to maintain stable keypoint density across scenes. 
Although descriptor learning relies on cosine similarity, matching is performed using Euclidean ($L_2$) distance, which is monotonically related for $L_2$-normalized vectors.

\subsection{On-Device Deployment}
\label{sec:mcu_method_validation}

To validate practical deployability, we perform end-to-end inference of \textit{Gideon} on STM32N6 with its embedded NPU.

Deployment proceeds in two stages. 
First, functional correctness and toolchain compatibility are verified on the STM32N6 Developer Kit using the ST Edge AI runtime. 
Second, quantitative latency and energy measurements are obtained on a dedicated STM32N657A platform under fixed-frequency operation (800\,MHz), representative of always-on scenarios.

Deployment is constrained by the 4.2\,MB on-chip SRAM budget, which must accommodate weights, activations, and runtime buffers. 
We explicitly verify that the compiled network executes fully within internal memory without external transfers, minimizing bandwidth overhead and latency variability. 
This requirement directly motivates compact feature maps and regular memory access patterns.

%% file: sec/3_results.tex
\section{Results and Discussion}
\label{sec:results}

\begin{figure}[t]
    \centering
    \includegraphics[width=\columnwidth]{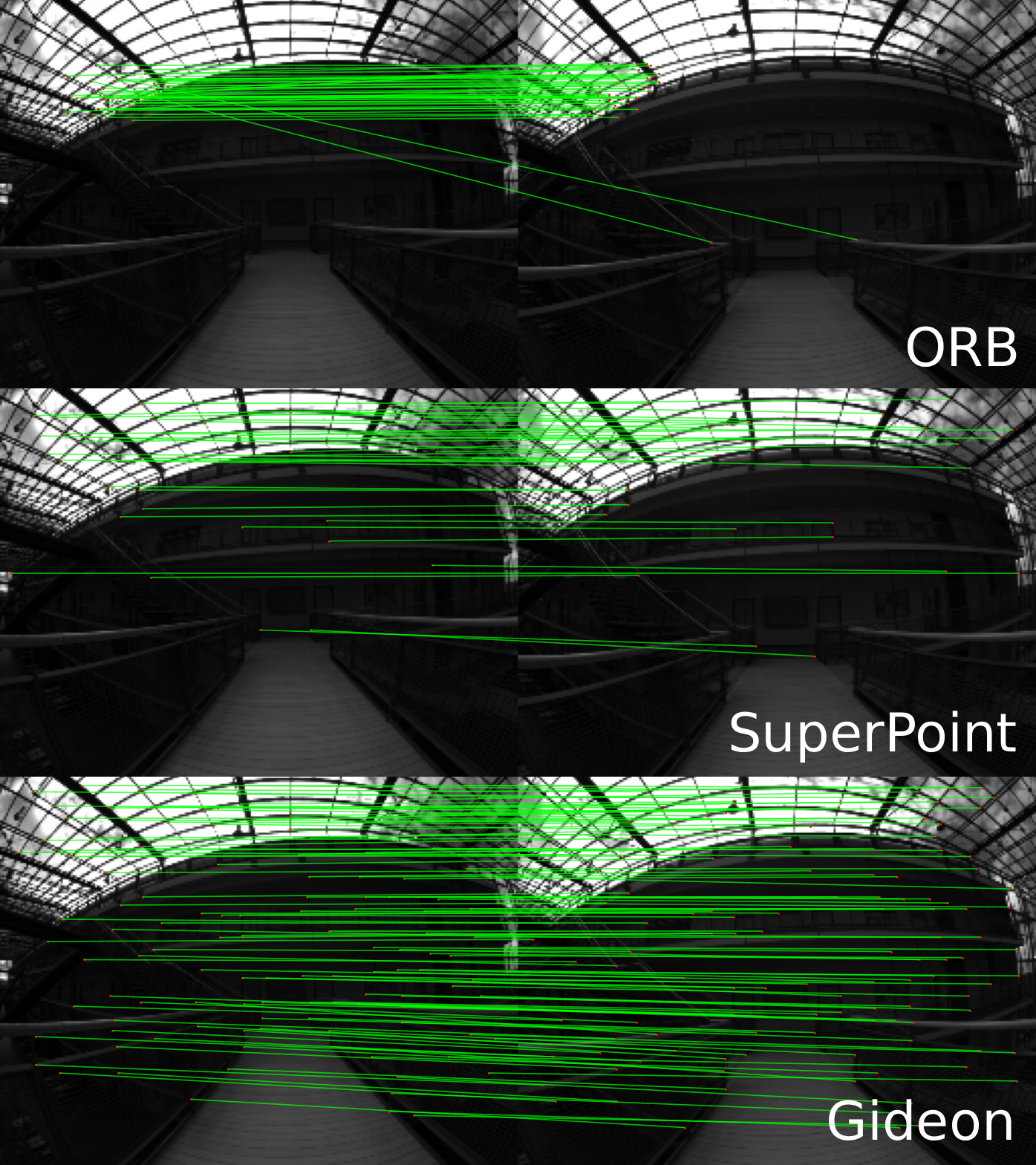}
    \caption{Qualitative results on TUM-VI. The green lines indicate predicted correspondences. The depicted image pair is not part of the training set.}
\label{fig:qualitative}
\end{figure}

\begin{table*}[tbp]
    \centering
    \caption{Comparison of Feature Extractors on Embedded Hardware using the HPatches dataset. Best results are highlighted in bold, computed separately for full-precision (float32) and quantized (int8) models.}
    \label{tab:model_comparison}
    \begin{tabular*}{\textwidth}{@{\extracolsep{\fill}}llcccccccc}
        \toprule
        \multirow{2}{*}{\textbf{Model}} 
        & \multirow{2}{*}{\textbf{Precision}} 
        & \multirow{2}{*}{\textbf{Time (ms)}} 
        & \multirow{2}{*}{\textbf{FPS}} 
        & \multirow{2}{*}{\textbf{\makecell{Weights \\ size}}}
        & \multirow{2}{*}{\textbf{\makecell{Activation \\ size}}}
        & \multicolumn{4}{c}{\textbf{HPatches Performance}} \\
        \cmidrule{7-10}
        & & & & & & \textbf{Rep (I)} & \textbf{Rep (V)} & \textbf{Cor (I)} & \textbf{Cor (V)} \\
        \midrule
        \multirow{2}{*}{\textbf{Gideon}} 
        & float32 & --- & --- & --- & --- & 0.594 & 0.478 & 0.919 & 0.593 \\
        & int8 & \textbf{9.003} & \textbf{111.07} & \textbf{600.77 KB} & \textbf{827.25 KB} & 0.574 & 0.474 & \textbf{0.937} & 0.614 \\
        \midrule
        \multirow{2}{*}{\textbf{SuperPoint}} 
        & float32 & --- & --- & --- & --- & \textbf{0.723} & \textbf{0.692} & \textbf{0.954} & \textbf{0.681} \\
        & int8 & 2108.604 & 0.47 & 1.24 MB & 4.61 MB & \textbf{0.720} & \textbf{0.693} & \textbf{0.937} & \textbf{0.668} \\
        \midrule
        \textbf{ORB} 
        & --- & --- & --- & --- & --- & 0.658 & 0.639 & 0.407 & 0.332 \\
        \bottomrule
    \end{tabular*}
\end{table*}

We evaluate Gideon along three axes: embedded runtime, quantization robustness, and descriptor dimensionality. To evaluate the performance and robustness of our proposed architecture, we conducted validation on the widely adopted HPatches dataset \cite{hpatches}, which is recognized as the standard benchmark for local feature evaluation.

HPatches provides a comprehensive suite of image sequences featuring severe illumination changes and drastic viewpoint variations. This environment perfectly mirrors the challenging conditions typically encountered in wearable SLAM applications and real-world hardware deployments.

As shown in Table \ref{tab:model_comparison}, Gideon achieves 9.003 ms inference time (111.07 fps) on the STM32N6 microcontroller and remains strictly under 1.5 MB for weights (600.78 kB) and activations (827.25 kB).

On the other hand, the proposed model achieves solid performance on HPatches, with an average repeatability of approximately 0.52, an illumination correctness exceeding 91\%, and a viewpoint correctness above 60\%.  It yields slightly lower metrics than SuperPoint, as expected in a distillation setup, but significantly surpasses the correctness of the standard ORB extractor.

Quantization induces a marginal degradation in keypoint localization (consistent with prior work on neural network quantization~\cite{8578384,9008784}), yet offers an unexpected improvement in descriptor matching performance. The average correctness of the INT8 descriptors surpasses that of the full-precision baseline, rising from 91.9\% to 93.7\% for illumination sequences and from 59.3\% to 61.4\% for viewpoint sequences. Although low-precision discretization has been reported to exhibit regularization-like effects in certain settings~\cite{binaryconnect,trainnoise}, these modest improvements may be partially attributed to statistical fluctuations. These improvements should be interpreted cautiously, as our training does not explicitly optimize for quantization. Further experiments would be needed to determine whether these effects are systematic.

As illustrated in Fig.~\ref{fig:qualitative}, Gideon yields a rich set of geometrically coherent matches across the scene, qualitatively complementing the quantitative improvements reported in Section~\ref{sec:results}. 

\subsection{Quantization as a Design Constraint}
Gideon’s architecture features a rapidly down-sampling stem followed by a heterogeneous mix of Inception-like parallel paths and Residual connections. Instead of more classical Batch Normalization layers, the NAS pipeline heavily favoured Affine layers, where scale and bias are treated as learnable parameters.

We conducted an ablation study evaluating four distinct network configurations to systematically isolate the contributions of specific architectural components, particularly concerning their impact on quantization resilience. The BatchNorm + ReLU represents a more conventional feature extraction design that works as the standard baseline. The BatchNorm + PWL (introducing piecewise-linear activations) and the Affine + ReLU (i.e. Gideon) configurations (replacing Batch Normalization with Affine layers) serve to independently assess the impact of our proposed modifications. Finally, we present Affine + PWL, which integrates both modifications.

\begin{table*}[tbp]
    \centering
    \caption{Relative performance change ($\Delta$\%) on the HPatches dataset due to int8 quantization. Positive values indicate an improvement after quantization.}
    \label{tab:quantization_loss}
    \begin{tabular*}{\textwidth}{@{\extracolsep{\fill}}lcccc}
        \toprule
        \multirow{2}{*}{\textbf{Architecture}} & \multicolumn{2}{c}{\textbf{$\Delta$ Detector Repeatability}} & \multicolumn{2}{c}{\textbf{$\Delta$ Descriptor Correctness}} \\
        \cmidrule(lr){2-3} \cmidrule(lr){4-5}
        & {\textbf{Illum.}} & {\textbf{View.}} & {\textbf{Illum.}} & {\textbf{View.}} \\
        \midrule
        \textbf{BatchNorm + ReLU} & -42.7\% & -51.1\% & -65.6\% & -100.0\% \\
        \textbf{BatchNorm + PWL}  & -27.8\% & -34.3\% & -32.3\% & -93.1\% \\
        \textbf{Affine + ReLU}    & -3.3\%  & -0.8\%  & \textbf{+1.9\%}  & \textbf{+3.4\%} \\
        \textbf{Affine + PWL}     & -2.3\%  & -5.2\%  & -6.4\%  & -30.9\% \\
        \midrule
        \textbf{SuperPoint}       & \textbf{-0.3\%}  & \textbf{+0.1\%}  & -1.8\%  & -2.0\% \\
        \bottomrule
    \end{tabular*}
\end{table*}

\subsubsection{BatchNorm Collapse}

The baseline Gideon architecture relying on BatchNorm and standard ReLU achieves the highest peak performance in full-precision. However, it suffers a near-total collapse when quantized to int8 as demonstrated in Table \ref{tab:quantization_loss}.

In fact, Batch Normalization produces a wide dynamic range of feature maps, significantly increasing the activation spread across channels. This severe cross-channel variance catastrophically amplifies quantization noise due to activation clipping, as a wider continuous range must be mapped onto a fixed, low-resolution discrete grid.

\subsubsection{Improving Quantization Resilience}

The proposed design works around a fundamental architectural trade-off between representational capacity and quantization readiness. In fact, Gideon sacrifices a marginal fraction of theoretical peak accuracy in the continuous domain to ensure stability in the discrete domain which has proven to be a necessary and highly advantageous compromise for edge deployment.

Similarly, the use of piecewise linear activation functions, such as Hardsigmoid and Hardtanh, improves the expressive power by increasing the information density in the activations, leading to slightly better quantization resilience.

Ultimately, replacing BatchNorm with Affine layers has proven to stabilize the network, effectively neutralizing post-training quantization degradation. As reported in Table \ref{tab:quantization_loss}, the quantized Affine + ReLU variant of Gideon not only maintains most of its original full-precision performance, but also slightly improves on a few metrics. While minor improvements are observed in some metrics after quantization, we note that our training does not employ quantization-aware optimization. Therefore, such variations may reflect statistical effects rather than a systematic regularization mechanism. Nevertheless, prior studies have reported regularization-like behavior induced by low-precision representations~\cite{binaryconnect,trainnoise}, which may provide a partial explanation.

\subsubsection{Adaptive Thresholding}

The proposed architectural variants of Gideon and the SuperPoint baseline were benchmarked under four thresholding conditions: three fixed confidence thresholds (0.005, 0.1, and 0.3) and one dynamic adaptive threshold as described in Section \ref{sec:detection_loss}, with consistent results.


A highly permissive threshold (0.005) maximizes Repeatability, approaching 80\% for illumination changes, but reduces Viewpoint Correctness to approximately 30--38\%. This is due to many weak, unstable keypoints producing low-quality matches that shift the consensus away from the correct solution.

Conversely, a restrictive threshold (0.3) acts as a strict stress test, extracting only a few, high-confidence points. Under these conditions, BatchNorm-based models collapse, whereas the Affine-based models demonstrate remarkable algorithmic robustness.

Nonetheless, the proposed Adaptive Threshold emerges as the optimal choice. Our tests show that it  acts as an intelligent filter: it removes noise while preserving structural features, thereby boosting Viewpoint Correctness to 60.3\% (up from 33.2\% at 0.005) without sacrificing overall Repeatability.

\subsection{Descriptor Dimensionality}

The loss formulated in Section \ref{sec:descriptor_loss} enables the distillation of the teacher's descriptors into representations of varying sizes. In this section, we investigate the impact of descriptor dimensionality on both representational capacity and quantization robustness through a comprehensive ablation study, scaling the channel depth from 8 to 512 components.

\subsubsection{Theoretical Analysis}
\label{sec:desc_dim_theory}

The standard deviation of L2-normalized descriptors decreases as their dimensionality increases. This behavior is mathematically expected: since the sum of the squared components must equal one, the average magnitude of each individual element must necessarily decrease proportionally to $1/\sqrt{D}$.

An analysis of the intra-frame distribution of descriptors shown in Table \ref{tab:descriptor_std} reveals that the representational space associated with lower dimensionalities (8, 16, and 32 channels) appears too constrained. This prevents the network from fully exploiting it, causing an information bottleneck that collapses the descriptors into lower-dimensional sub-manifolds.

Conversely, for dimensions of 64 and above, the ratio between the empirical standard deviation and the ideal theoretical value stabilizes at approximately 0.75. This indicates that, from 64 channels onward, the information is distributed in an almost perfectly isotropic manner (i.e. uniformly across all directions within the hyperspherical space).

Finally, high-dimensional descriptors, such as those with 128 (e.g. SIFT\cite{Lowe:2004:DIF:993451.996342}), 256 (e.g. SuperPoint) or 512 floating-point components, exhibit a negligible standard deviation (0.047 and 0.033, respectively). This occurs because the majority of the values generated by the projection layer oscillate very close to zero. This narrow dynamic range precludes effective INT8 quantization as a significant portion of the representational expressiveness is lost due to the limited resolution of the discrete grid, causing this setup to become highly suboptimal for edge hardware deployments.

\subsubsection{Empirical Validation}

Empirical results on the HPatches dataset, reported in Table \ref{tab:desc_dim_ablation}, strongly corroborate our theoretical analysis regarding intra-frame standard deviation presented in Section \ref{sec:desc_dim_theory}.

At lower dimensions, we observe a classical under-parameterization phenomenon. For instance, with an 8-dimensional representation, performance is severely degraded, yielding a Viewpoint Correctness of approximately 0.24. Scaling to 16 dimensions provides a substantial performance leap, with Illumination Correctness increasing from 0.64 to 0.84, although viewpoint-related metrics remain suboptimal. At these lower dimensions, the representational vector is fundamentally too compressed to effectively encode the geometric and photometric complexity of a local patch.

Interestingly, expanding the capacity to 32 dimensions reveals a counterintuitive behavior where the Viewpoint Correctness in INT8 (0.613) surpasses that of its FP32 counterpart (0.583). The INT8 variant occasionally matches or slightly surpasses the full-precision model, probably due to statistical variability.

Building upon these observations, our ablation study demonstrates that constraining the network to 64 channels acts as a highly effective structural regularizer, yielding the ideal equilibrium for edge hardware. At this dimensionality, the learned representation carries sufficient information distributed isotropically across its vector space. This forces the projection layer to output highly discriminative features, achieving an optimal convergence between representational capacity and quantization robustness. In fact, floating-point performance stabilizes at excellent values (0.922 for Illumination and 0.613 for Viewpoint), while the INT8 version closely matches it with negligible degradation, proving that the dynamic range at 64 dimensions remains broad enough to robustly withstand 8-bit discretization.

In contrast, at 256 or 512 dimensions, the network exhibits capacity underutilization, populating many channels with near-zero values. This implicit sparsity does not improve matching robustness but increases memory and latency.

Even though increasing the dimensionality continues to yield marginal theoretical benefits in the floating-point domain, with Viewpoint Correctness slowly climbing to a peak of 0.654 at 512 dimensions, this expansion compresses the activation values closer to zero, resulting in a shrinking standard deviation. Consequently, this constrained dynamic range precludes effective INT8 quantization, as the narrow distribution requires severe approximation to be mapped onto the 256 discrete bins of the INT8 format. Indeed, at 512 dimensions, the INT8 network achieves a lower Viewpoint Correctness (0.586) than it does with merely 32 dimensions (0.613).

Ultimately, a 64-channel descriptor emerges as the optimal architectural choice. Its broader dynamic range ensures that, during INT8 quantization, information is distributed across a significantly higher number of discrete bins, thereby fully preserving the discriminative capacity of the descriptor when deployed on the edge.

\subsection{On-Device Energy Evaluation}
\label{sec:mcu_results}


Peak runtime validation was performed on the STM32N6 Developer Kit (ST Edge AI v3.0.0) at 1\,GHz, yielding 9.003\,ms latency (111.07\,fps). For power characterization, we evaluated the same compiled model on a dedicated STM32N657A platform under fixed-frequency operation (800\,MHz).

End-to-end inference completes in 10.87\,ms with an average current of 274.6\,mA, corresponding to $Q=2.984$\,mC and $E_{\mathrm{inf}}=5.372$\,mJ per inference at 1.8\,V. 
At 60\,fps continuous operation, this maps to an inference-only power of approximately 322\,mW.

From a memory perspective, the INT8 deployment requires 600.77\,KB of weights and 827.25\,KB peak activations, remaining well within the 4.2\,MB on-chip SRAM budget and avoiding external memory transfers.

\begin{table}[t]
    \centering
    \caption{Analysis of intra-frame standard deviation across different descriptor dimensions. The theoretical standard deviation is calculated as $1/\sqrt{D}$, assuming a perfectly isotropic distribution on the L2-normalized hypersphere. 
    }
    \label{tab:descriptor_std}
    \begin{tabular*}{\columnwidth}{@{\extracolsep{\fill}}lccc@{}}
        \toprule
        \textbf{Dim.} & \textbf{Theor. Std.} & \textbf{Meas. Std.} & \textbf{Ratio} \\
        \midrule
        8   & 0.3536 & 0.1701 & 0.4811 \\
        16  & 0.2500 & 0.1381 & 0.5526 \\
        32  & 0.1768 & 0.1139 & 0.6443 \\
        64  & 0.1250 & 0.0899 & 0.7188 \\
        128 & 0.0884 & 0.0672 & 0.7606 \\
        256 & 0.0625 & 0.0472 & 0.7545 \\
        512 & 0.0442 & 0.0332 & 0.7510 \\
        \bottomrule
    \end{tabular*}
\end{table}

\begin{table}[t]
    \centering
    \caption{Ablation study on descriptor dimensionality evaluated on the HPatches dataset. Values in parentheses show the relative change ($\Delta$\%) due to quantization.}
    \label{tab:desc_dim_ablation}
    \begin{tabular*}{\columnwidth}{@{\extracolsep{\fill}}llcc}
        \toprule
        & & \multicolumn{2}{c}{\textbf{Descriptor Correctness}} \\
        \cmidrule(lr){3-4}
        \textbf{Dim ($D$)} & \textbf{Precision} & {\textbf{Illum.}} & {\textbf{View.}} \\
        \midrule
        
        \multirow{2}{*}{\textbf{8}} 
        & float32 & 0.6421 & 0.2475 \\
        & int8    & 0.6316 {\scriptsize (-1.6\%)} & 0.2339 {\scriptsize (-5.5\%)} \\
        \midrule
        
        \multirow{2}{*}{\textbf{16}}  
        & float32 & 0.8491 & 0.5085 \\
        & int8    & 0.8596 {\scriptsize (+1.2\%)} & 0.5186 {\scriptsize (+2.0\%)} \\
        \midrule
        
        \multirow{2}{*}{\textbf{32}}    
        & float32 & 0.9123 & 0.5831 \\
        & int8    & 0.8947 {\scriptsize (-1.9\%)} & \textbf{0.6136} {\scriptsize (+5.2\%)} \\
        \midrule

        \multirow{2}{*}{\textbf{64}}     
        & float32 & 0.9193 & 0.5932 \\
        & int8    & \textbf{0.9368} {\scriptsize (+1.9\%)} & \textbf{0.6136} {\scriptsize (+3.4\%)} \\
        \midrule

        \multirow{2}{*}{\textbf{128}}     
        & float32 & 0.9404 & 0.6136 \\
        & int8    & 0.9053 {\scriptsize (-3.7\%)} & 0.6068 {\scriptsize (-1.1\%)} \\
        \midrule

        \multirow{2}{*}{\textbf{256}}     
        & float32 & \textbf{0.9439} & 0.6339 \\
        & int8    & 0.9193 {\scriptsize (-2.6\%)} & 0.6000 {\scriptsize (-5.3\%)} \\
        \midrule

        \multirow{2}{*}{\textbf{512}}     
        & float32 & 0.9298 & \textbf{0.6542} \\
        & int8    & 0.9158 {\scriptsize (-1.5\%)} & 0.5864 {\scriptsize (-10.4\%)} \\
        
        \bottomrule
    \end{tabular*}
\end{table}

%% file: sec/4_conclusion.tex
\section{Conclusions}

In this work we presented \textit{Gideon}, a hardware-aware neural feature extractor designed for deployment on microcontroller-class devices. 
Rather than optimizing solely for theoretical efficiency, we re-framed feature extraction as a system-level problem, jointly considering memory footprint, dataflow regularity, and quantization stability as first-class design constraints.

By combining relational knowledge distillation with differentiable architecture search under embedded constraints, we obtained a compact convolutional topology that preserves the structural geometry of a strong teacher model while remaining fully deployable on STM32N6-class hardware.

Our experiments reveal that quantization robustness is largely architectural: replacing Batch Normalization with Affine layers and controlling descriptor dimensionality substantially improves INT8 stability. 
With 9\,ms latency and a total footprint below 1.5\,MB, Gideon demonstrates that learned local features can operate reliably in always-on embedded scenarios. 
While full SLAM integration is beyond the scope of this work, we report feature-level metrics such as repeatability and matching correctness, which are widely accepted indicators of downstream SLAM performance. The strong results obtained on HPatches therefore suggest promising applicability within complete SLAM pipelines.

Ultimately, the key contribution of this work lies in framing feature extraction as a system-level co-design problem, where memory constraints, dataflow regularity, and quantization stability are explicitly integrated into the training process, rather than addressed only at deployment time. 
This perspective enables the development of robust and efficient perception modules tailored for next-generation wearable devices, aligning with broader trends in efficient neural network design for resource-constrained environments~\cite{howard2017mobilenets,pmlr-v97-tan19a}.


%% file: main.bib
@String(CVPR= {IEEE Conf. Comput. Vis. Pattern Recog.})

@String(ICCV= {Int. Conf. Comput. Vis.})

@String(ECCV= {Eur. Conf. Comput. Vis.})

@String(NIPS= {Adv. Neural Inform. Process. Syst.})

@String(ICLR = {Int. Conf. Learn. Represent.})

@String(CVPR  = {CVPR})

@String(ICCV  = {ICCV})

@String(ECCV  = {ECCV})

@String(NIPS  = {NeurIPS})

@String(ICLR  = {ICLR})

@article{orbslam3,
   title={ORB-SLAM3: An Accurate Open-Source Library for Visual, Visual–Inertial, and Multimap SLAM},
   volume={37},
   ISSN={1941-0468},
   url={http://dx.doi.org/10.1109/TRO.2021.3075644},
   DOI={10.1109/tro.2021.3075644},
   number={6},
   journal={IEEE Transactions on Robotics},
   publisher={Institute of Electrical and Electronics Engineers (IEEE)},
   author={Campos, Carlos and Elvira, Richard and Rodriguez, Juan J. Gomez and M. Montiel, Jose M. and D. Tardos, Juan},
   year={2021},
   month=dec, pages={1874–1890}
}

@INPROCEEDINGS{orb,
  author={Rublee, Ethan and Rabaud, Vincent and Konolige, Kurt and Bradski, Gary},
  booktitle={2011 International Conference on Computer Vision}, 
  title={ORB: An efficient alternative to SIFT or SURF}, 
  year={2011},
  volume={},
  number={},
  pages={2564-2571},
  doi={10.1109/ICCV.2011.6126544}
}

@inproceedings{fast,
  author = {Rosten, Edward and Drummond, Tom},
  title = {Machine Learning for High-Speed Corner Detection},
  booktitle = {ECCV},
  series = {LNCS},
  volume = {3951},
  pages = {430--443},
  year = {2006},
  publisher = {Springer}
}

@INPROCEEDINGS{brisk,
  author={Leutenegger, Stefan and Chli, Margarita and Siegwart, Roland Y.},
  booktitle={2011 International Conference on Computer Vision}, 
  title={BRISK: Binary Robust invariant scalable keypoints}, 
  year={2011},
  volume={},
  number={},
  pages={2548-2555},
  doi={10.1109/ICCV.2011.6126542}
}

@article{superpoint,
  title={SuperPoint: Self-Supervised Interest Point Detection and Description},
  author={DeTone, Daniel and Malisiewicz, Tomasz and Rabinovich, Andrew},
  journal={arXiv preprint arXiv:1712.07629},
  year={2018}
}

@inproceedings{r2d2,
  author = {Revaud, Jerome and De Souza, Cesar and Humenberger, Martin and Weinzaepfel, Philippe},
  booktitle = {Advances in Neural Information Processing Systems},
  editor = {H. Wallach and H. Larochelle and A. Beygelzimer and F. d\textquotesingle Alch\'{e}-Buc and E. Fox and R. Garnett},
  pages = {},
  publisher = {Curran Associates, Inc.},
  title = {R2D2: Reliable and Repeatable Detector and Descriptor},
  url ={https://proceedings.neurips.cc/paper_files/paper/2019/file/3198dfd0aef271d22f7bcddd6f12f5cb-Paper.pdf},
  volume = {32},
  year = {2019}
}

@inproceedings{loftr,
  title={LoFTR: Detector-Free Local Feature Matching with Transformers},
  author={Sun, Jiaming and Shen, Zehong and Wang, Yuang and Bao, Hujun and Zhou, Xiaowei},
  booktitle={CVPR},
  year={2021}
}

@misc{xfeat,
  title={XFeat: Accelerated Features for Lightweight Image Matching}, 
  author={Guilherme Potje and Felipe Cadar and Andre Araujo and Renato Martins and Erickson R. Nascimento},
  year={2024},
  eprint={2404.19174},
  archivePrefix={arXiv},
  primaryClass={cs.CV},
  url={https://arxiv.org/abs/2404.19174}, 
}

@misc{aliked,
  title={ALIKED: A Lighter Keypoint and Descriptor Extraction Network via Deformable Transformation}, 
  author={Xiaoming Zhao and Xingming Wu and Weihai Chen and Peter C. Y. Chen and Qingsong Xu and Zhengguo Li},
  year={2023},
  eprint={2304.03608},
  archivePrefix={arXiv},
  primaryClass={cs.CV},
  url={https://arxiv.org/abs/2304.03608}, 
}

@inproceedings{darts,
  title={DARTS: Differentiable Architecture Search},
  author={Liu, Hanxiao and Simonyan, Karen and Yang, Yiming},
  booktitle={ICLR},
  year={2019}
}

@inproceedings{gs1,
  title={Categorical Reparameterization with Gumbel-Softmax},
  author={Jang, Eric and Gu, Shixiang and Poole, Ben},
  booktitle={ICLR},
  year={2017}
}

@inproceedings{gs2,
  title={The Concrete Distribution: A Continuous Relaxation of Discrete Random Variables},
  author={Maddison, Chris J. and Mnih, Andriy and Teh, Yee Whye},
  booktitle={ICLR},
  year={2017}
}

@inproceedings{tumvi,
  title={The TUM VI Benchmark for Evaluating Visual-Inertial Odometry},
  author={Schubert, David and Goll, Thomas and Demmel, Nikolaus and Usenko, Vladyslav and St{\"u}ckler, J{\"o}rg and Cremers, Daniel},
  booktitle={IEEE/RSJ International Conference on Intelligent Robots and Systems (IROS)},
  year={2018}
}

@misc{uncertaintyweighting,
    title={Multi-Task Learning Using Uncertainty to Weigh Losses for Scene Geometry and Semantics}, 
    author={Alex Kendall and Yarin Gal and Roberto Cipolla},
    year={2018},
    eprint={1705.07115},
    archivePrefix={arXiv},
    primaryClass={cs.CV},
    url={https://arxiv.org/abs/1705.07115}, 
}

@article{kullback1951information,
  title={On information and sufficiency},
  author={Kullback, Solomon and Leibler, Richard A},
  journal={Annals of Mathematical Statistics},
  year={1951}
}

@inproceedings{hinton2015distilling,
  title={Distilling the Knowledge in a Neural Network},
  author={Hinton, Geoffrey and Vinyals, Oriol and Dean, Jeff},
  booktitle={NIPS Deep Learning Workshop},
  year={2015}
}

@inproceedings{hpatches,
  title={HPatches: A benchmark and evaluation of handcrafted and learned local descriptors},
  author={Balntas, Vassileios and Lenc, Karel and Vedaldi, Andrea and Mikolajczyk, Krystian},
  booktitle={CVPR},
  year={2017}
}

@misc{binaryconnect,
  title={BinaryConnect: Training Deep Neural Networks with binary weights during propagations}, 
  author={Matthieu Courbariaux and Yoshua Bengio and Jean-Pierre David},
  year={2016},
  eprint={1511.00363},
  archivePrefix={arXiv},
  primaryClass={cs.LG},
  url={https://arxiv.org/abs/1511.00363}, 
}

@InProceedings{pmlr-v97-tan19a,
  title = 	 {{E}fficient{N}et: Rethinking Model Scaling for Convolutional Neural Networks},
  author =       {Tan, Mingxing and Le, Quoc},
  booktitle = 	 {Proceedings of the 36th International Conference on Machine Learning},
  pages = 	 {6105--6114},
  year = 	 {2019},
  editor = 	 {Chaudhuri, Kamalika and Salakhutdinov, Ruslan},
  volume = 	 {97},
  series = 	 {Proceedings of Machine Learning Research},
  month = 	 {09--15 Jun},
  publisher =    {PMLR},
  url = 	 {https://proceedings.mlr.press/v97/tan19a.html}
}

@misc{howard2017mobilenets,
  author = {Howard, Andrew G. and Zhu, Menglong and Chen, Bo and Kalenichenko, Dmitry and Wang, Weijun and Weyand, Tobias and Andreetto, Marco and Adam, Hartwig},
  description = {MobileNets: Efficient Convolutional Neural Networks for Mobile Vision
  Applications},
  note = {cite arxiv:1704.04861},
  title = {MobileNets: Efficient Convolutional Neural Networks for Mobile Vision
  Applications},
  url = {http://arxiv.org/abs/1704.04861},
  year = 2017
}

@article{Lowe:2004:DIF:993451.996342,
  acmid = {996342},
  address = {Hingham, MA, USA},
  author = {Lowe, David G.},
  description = {Distinctive Image Features from Scale-Invariant Keypoints},
  doi = {10.1023/B:VISI.0000029664.99615.94},
  issn = {0920-5691},
  issue_date = {November 2004},
  journal = {Int. J. Comput. Vision},
  month = nov,
  number = 2,
  numpages = {20},
  pages = {91--110},
  publisher = {Kluwer Academic Publishers},
  title = {Distinctive Image Features from Scale-Invariant Keypoints},
  url = {http://dx.doi.org/10.1023/B:VISI.0000029664.99615.94},
  volume = 60,
  year = 2004
}

@INPROCEEDINGS{9008784,
  author={Nagel, Markus and Baalen, Mart Van and Blankevoort, Tijmen and Welling, Max},
  booktitle={2019 IEEE/CVF International Conference on Computer Vision (ICCV)}, 
  title={Data-Free Quantization Through Weight Equalization and Bias Correction}, 
  year={2019},
  volume={},
  number={},
  pages={1325-1334},
  doi={10.1109/ICCV.2019.00141}}

@INPROCEEDINGS{8578384,
  author={Jacob, Benoit and Kligys, Skirmantas and Chen, Bo and Zhu, Menglong and Tang, Matthew and Howard, Andrew and Adam, Hartwig and Kalenichenko, Dmitry},
  booktitle={2018 IEEE/CVF Conference on Computer Vision and Pattern Recognition}, 
  title={Quantization and Training of Neural Networks for Efficient Integer-Arithmetic-Only Inference}, 
  year={2018},
  volume={},
  number={},
  pages={2704-2713},
  doi={10.1109/CVPR.2018.00286}}

@inproceedings{DBLP:journals/corr/RomeroBKCGB14,
  author       = {Adriana Romero and
                  Nicolas Ballas and
                  Samira Ebrahimi Kahou and
                  Antoine Chassang and
                  Carlo Gatta and
                  Yoshua Bengio},
  editor       = {Yoshua Bengio and
                  Yann LeCun},
  title        = {FitNets: Hints for Thin Deep Nets},
  booktitle    = {3rd International Conference on Learning Representations, {ICLR} 2015,
                  San Diego, CA, USA, May 7-9, 2015, Conference Track Proceedings},
  year         = {2015},
  url          = {http://arxiv.org/abs/1412.6550},
  bibsource    = {dblp computer science bibliography, https://dblp.org}
}

@inproceedings{lindenberger2023lightglue,
  author    = {Philipp Lindenberger and
               Paul-Edouard Sarlin and
               Marc Pollefeys},
  title     = {{LightGlue: Local Feature Matching at Light Speed}},
  booktitle = {ICCV},
  year      = {2023}
}

@inproceedings{NEURIPS2020_a42a596f,
 author = {Tyszkiewicz, Micha\l  and Fua, Pascal and Trulls, Eduard},
 booktitle = {Advances in Neural Information Processing Systems},
 editor = {H. Larochelle and M. Ranzato and R. Hadsell and M.F. Balcan and H. Lin},
 pages = {14254--14265},
 publisher = {Curran Associates, Inc.},
 title = {DISK: Learning local features with policy gradient},
 url = {https://proceedings.neurips.cc/paper_files/paper/2020/file/a42a596fc71e17828440030074d15e74-Paper.pdf},
 volume = {33},
 year = {2020}
}

@InProceedings{Dusmanu2019CVPR,
    author = {Dusmanu, Mihai and Rocco, Ignacio and Pajdla, Tomas and Pollefeys, Marc and Sivic, Josef and Torii, Akihiko and Sattler, Torsten},
    title = {{D2-Net: A Trainable CNN for Joint Detection and Description of Local Features}},
    booktitle = {Proceedings of the 2019 IEEE/CVF Conference on Computer Vision and Pattern Recognition},
    year = {2019},
}

@misc{trainnoise,
  title={Training with Quantization Noise for Extreme Model Compression}, 
  author={Angela Fan and Pierre Stock and Benjamin Graham and Edouard Grave and Remi Gribonval and Herve Jegou and Armand Joulin},
  year={2021},
  eprint={2004.07320},
  archivePrefix={arXiv},
  primaryClass={cs.LG},
  url={https://arxiv.org/abs/2004.07320}, 
}

@article{nanoSLAM2309.12008,
  title={NanoSLAM: Enabling Fully Onboard SLAM for Tiny Robots},
  author={Niculescu, Vlad and Polonelli, Tommaso and Magno, Michele and Benini, Luca},
  journal={IEEE Internet of Things Journal},
  year={2023},
  publisher={IEEE}
}

@misc{levio2602.03294,
  title={LEVIO: Lightweight Embedded Visual-Inertial Odometry for Resource-Constrained Devices},
  author={Kühne et al.},
  year={2026},
  eprint={2602.03294},
  archivePrefix={arXiv},
  primaryClass={cs.CV}
}
